\documentclass{article}

\PassOptionsToPackage{numbers, compress}{natbib}

\usepackage[final]{nips_2017}
\pdfoutput=1

\usepackage[utf8]{inputenc} 
\usepackage[T1]{fontenc}    
\usepackage{hyperref}       
\usepackage{url}            
\usepackage{booktabs}       
\usepackage{amsfonts}       
\usepackage{nicefrac}       
\usepackage{microtype}      
\usepackage{bm}
\usepackage{nomencl}
\usepackage{pdfpages}
\usepackage{amssymb}
\usepackage{mathrsfs}
\usepackage{amsthm}
\usepackage[cmex10]{amsmath}
\usepackage{caption}
\usepackage{dsfont}

\usepackage{algorithm}
\usepackage[noend]{algpseudocode}

\def\C{\bm{C}}

\def\S{\bm{S}}

\def\X{\bm{X}}

\def\Z{\bm{Z}}

\def\f{\bm{f}}
\def\g{\bm{g}}

\def\x{\bm{x}}

\def\z{\bm{z}}

\def\btheta{\boldsymbol{\theta}}

\def\bmu{\boldsymbol{\mu}}

\def\argmin#1{\underset{#1}{\textrm{argmin}}}
\def\argmax#1{\underset{#1}{\textrm{argmax}}}
\def\maxim#1{\underset{#1}{\textrm{max}}}

\def\minim#1{\underset{#1}{\textrm{min}}}

\def\setD{\mathcal{D}}
\def\cardD{|\mathcal{D}|}
\def\setB{\mathcal{B}}
\def\cardB{|\mathcal{B}|}
\def\setA{\mathcal{A}}
\def\cardA{|\mathcal{A}|}
\def\setCk{\mathcal{C}_k}
\def\setCl{\mathcal{C}_\ell}
\def\cardCk{|\mathcal{C}_k|}
\def\cardCl{|\mathcal{C}_\ell|}
\def\Nw{\mathcal{N}_w}
\def\Nb{\mathcal{N}_b}
\def\nrmz{\tilde{\z}}
\def\centvars{\{\bmu_k\}_{k=1}^K}

\title{Deep Discriminative Latent Space for Clustering }

\author{
  Elad~Tzoreff\thanks{Both authors contributed equally} \\
  Huawei Tel Aviv Research Center\\
  Hod Hasharon\\
  \texttt{elad.tzoreff@huawei.com} \\
\And
Olga~Kogan\footnotemark[1]\\
Huawei Tel Aviv Research Center\\
Hod Hasharon\\
\texttt{olga.kogan@huawei.com}
\And
Yoni~Choukroun\\
Huawei Tel Aviv Research Center\\
Hod Hasharon\\
\texttt{yoni.choukroun@huawei.com} \\
}
\begin{document}

\maketitle

\begin{abstract}
\textbf{Clustering is one of the most fundamental tasks in data analysis and machine learning. It is central to many data-driven applications that aim to separate the data into groups with similar patterns. Moreover, clustering is a complex procedure that is affected significantly by the choice of the data representation method. Recent research has demonstrated encouraging clustering results by learning effectively these representations. In most of these works a deep auto-encoder is initially pre-trained to minimize a reconstruction loss, and then jointly optimized with clustering centroids in order to improve the clustering objective. Those works focus mainly on the clustering phase of the procedure, while not utilizing the potential benefit out of the initial phase. In this paper we propose to optimize an auto-encoder with respect to a discriminative pairwise loss function during the auto-encoder pre-training phase. We demonstrate the high accuracy obtained by the proposed method as well as its rapid convergence (e.g. reaching above $92\%$ accuracy on MNIST during the pre-training phase, in less than 50 epochs), even with small networks.
}

\end{abstract}

\section{Introduction}Traditionally, most learning approaches have been treating representation-learning/feature-selection and clustering separately. However recent studies have outperformed traditional methods by learning optimal representations for clustering. In most of these works a deep auto-encoder is first trained to reduce a reconstruction loss. Next,  the encoder parameters and clustering parameters (e.g. the K-means centroids)  are jointly optimized in order to improve the overall clustering accuracy. However, we observed that in most cases the improvement of the clustering phase over the pre-training phase amounts no more than $15-20\%$ percents of accuracy. Therefore, reaching a high level of accuracy in the pre-raining phase is of crucial importance. Moreover, a reconstruction loss is not an optimal choice for clustering, due to the natural trade-off between reconstruction and clustering. A reconstruction aims to reproduce every detail in the original data, while clustering aims to reduce all possible variations into several templates. 

In this paper we propose a novel unified framework for learning a clustering oriented representation. We suggest the optimization of an auto-encoder in the pre-training phase, with respect to a discriminative loss function which encourage a clustering oriented representation. The discriminative loss is the weighted sum of all pairwise similarities among data-points in the batch. Minimization of this loss implies making data-points' representation as dissimilar as possible. Under the assumption of a balanced dataset, the majority of the pairs are indeed dissimilar and only a small fraction of them are similar (i.e., only the within cluster pairs). Accordingly, the utilization of this loss is justified. The proposed optimization scheme enables the utilization of relatively small networks whom can be trained very fast. For the clustering phase we propose a joint optimization scheme that maintain the pairwise discrimination loss while it optimizes the clustering objective. We apply the proposed algorithm to several datasets  (MNIST, COIL-20, COIL-100), and demonstrate its superiority both in terms of accuracy and speed of convergence. To summarize, the contributions of this paper are twofold: (1) the major contribution of this paper is a novel optimization scheme for the auto-encoder pre-training phase, that encourages a discriminative latent space which fits with the clustering objective and reaches higher accuracy prior to the clustering phase, and (2) a novel clustering scheme in which the discrimination of the latent space is strengthened while searching for the optimal centroids. 

\section{Related work}
We provide here a brief review of the currently known methods that optimize the latent space for clustering.  \citep{xie2016unsupervised} proposed DEC - a fully connected stacked auto encoder that learns a latent representation by minimizing the reconstruction loss in the pre-training phase.  The objective function applied to the clustering phase, is the Kullback Leibler (KL) divergence between the clustering soft assignments modeled as a t-distribution, and a reference heuristic probability that is constructed from the soft assignments. In \citep{yang2016towards}  a fully connected auto-encoder is trained with a k-means loss along with a reconstruction loss at the clustering phase. \citep{li2017discriminatively} proposed DBC - a fully convolutional  network with layer-wised batch normalization that aims to overcome the relatively slow training of a stacked auto-encoder.  DBC utilized the same objective function as DEC with a boosting factor as a hyperparameter, for the reference probability. \citep{yang2016joint} introduced JULE -  a method that jointly optimize a convolutional neural network with the clustering parameters in a recurrent manner using an agglomerative clustering approach. In \citep{jiang2016variational} VaDE  - a variational auto encoder for deep embedding is proposed. A generative model from which a cluster is picked and a latent representation is selected followed by a deep network which decodes the latent embedding into the observables. In \citep{ji2017deep} a deep auto-encoder is trained to minimize a reconstruction loss together with a self expressive layer.  This objective encourages a sparse representation of the original data. \citep{dizaji2017deep} proposed DEPICT - a method that train a convolutional auto-encoder with a softmax layer stacked on-top of the encoder. The softmax entries represent the assignment of each data-point to each one of the clusters. Finally, following \citep{shah2017robust}, a deep continuous clustering approach is suggested in \citep{shah2018deep}. In this method the auto-encoder parameters are simultaneously optimized together with a set of representatives defined against each data-point. The representatives are optimized by minimizing the distance between each representative and its associated data-point, while minimizing the pairwise distances between representatives (similar to the convex clustering approach proposed by \citep{chi2015splitting}). They applied non-convex objective functions to penalize for the pairwise distances between the representatives. The clusters are then determined as the connected components in the graph created between the representatives.

\section{Discriminative clustering}
\label{prob_stat}
In this paper we choose the paradigm in which a latent space representation of a certain dataset is trained jointly with clustering parameters. This end-to-end approach, backed by recent papers, achieves a clustering oriented representation and therefore better clustering accuracy. In most papers the auto- encoder is first trained to minimize the reconstruction loss and then applied as an initial condition to the joint optimization problem in which both clustering and encoder parameters are jointly optimized to minimize the clustering error. However although a major attention is dedicated to the clustering phase, we have observed that in most cases the improvement of the clustering phase over the initial phase amounts a maximum of $15-20\%$ percents of accuracy. This leads to the conclusion that the initial phase has a significant effect on the overall accuracy and therefore a focus shall be put on this step. 

\subsection{The pre-training phase: obtaining discriminative latent space}
Let $\mathcal{D}$  denotes a dataset grouped into $K$ clusters $\mathcal\{C_k\}_{k=1}^K$, and let $\x_i \in R^p$ be a data-point in the set with dimension $p$. Let $\z_i = \f(\x_i; \btheta_e) \in R^d $ stands for the latent representation of $\x_i$, where the parameters of the encoder are denoted by $\btheta_e$, and $p \gg d$. Let $\hat{\x}_i = \g(\z_i; \btheta_d)$ stands for the reconstructed data point, that is, the output of the decoder, where the parameters of the decoder are denoted by $\btheta_d$.  

We propose a family of pairwise discriminative functions of the form $L_d(\z_i, \z_j): R^{d \times d} \rightarrow R$, where $L_d(\z_i, \z_j) = \text{sim}(\z_i, \z_j)$, and $\text{sim}(\cdot)$ stands for any similarity measure between a pair of data-points. 
\begin{align}
\label{eq_discrim_family}
L_d(\setD; \btheta_e) = \sum_{i, j \in \mathcal{D}} w_{ij}\text{sim}(\z_i, \z_j)
\end{align}
where $w_{ij}$ are related to the similarities between in the raw data-points. When prior knowledge is not available we set  $w_{i,j} = |\mathcal{D}| ^{-2}$ , where $\cardD$ stands for the cardinality of the dataset.
Note that the objective function in \eqref{eq_discrim_family} with $w_{i,j} = |\mathcal{D}| ^{-2}$ is sub-optimal, since it penalizes all similarities regardless of whether they belong to the same cluster of not. Obviously if the assignments of each data point were available, $w_{i,j}$ have been split into two sets of weights: one for the minimization of the cross clusters similarities,  and one for the maximization of within cluster similarities, yielding the following objective function instead:
\begin{align}
\label{eq_optimal}
L_d(\setD; \btheta_e) =\frac{1}{\mathcal{N}_b } \sum_{i, j \notin \mathcal{C}} \text{sim}(\z_i, \z_j)  -  \frac{1}{\mathcal{N}_w} \sum_{i, j \in \mathcal{C}} \text{sim}(\z_i, \z_j)
\end{align}
where the notation $i,j \in \mathcal{C}$ defines a pair of data-points $i,j$ related to the same cluster $\mathcal{
C}$. $\mathcal{N}_b , \mathcal{N}_w$ stands for the number of between cluster and within cluster pairs, respectively. However, the justification of \eqref{eq_discrim_family} with $w_{i,j} = |\mathcal{D}| ^{-2}$ arises from the following observation, consider a balanced dataset that is not dominated by a single or a few clusters, then  the number of pairs in $\mathcal{D}$ are $\binom{\cardD}{2}$ and the within cluster and between cluster pairs cardinalities are approximately $\Nw \sim K\,\binom{\cardD/K}{2}$ and $\Nb \sim\frac{1}{2}\cardD^2(1 - \frac{1}{K})$, respectively. Accordingly, there are much more dissimilar pairs than similar ones. Note that $\Nb$ increases with both $\cardD, K$, and for $\cardD \gg K$, the number of within cluster pairs  is approximately $\frac{1}{K}$ fraction of all data pairs.

In light of eq. \eqref{eq_optimal} we create a k-nearest neighbors graph between the data-points based on their original representation. A fraction of pairs with the largest similarities from the k-nearest neighbors graph is then applied to eq. \eqref{eq_discrim_family} as anchor pairs, whom similarity is to be maximized. The maximization of anchors' similarities refers to the within cluster component in eq. \eqref{eq_optimal}. Utilization of only a small fraction of pairs from the k-nearest neighbors graph is motivated by the need to avoid maximization of similarities between pairs that are actually dissimilar. Since similarities based on the original representation are not reliable we only use the pairs with the highest confidence.  Define $\mathcal{A}$ as the set of anchor pairs, we have
\begin{align}
\label{eq_wij}
w_{ij} = \left\{  \begin{array}{l l}
- \frac{1-\alpha}{\cardA} &\,\,  i,j, \in \mathcal{A}\\
\frac{1}{\cardD^{2} - \cardA} & \,\,  i,j, \notin \mathcal{A} 
\end{array} \right.\
\end{align}
yields
\begin{align}
\label{eq_discrim}
L_d(\setD; \btheta_e) = \frac{1}{\cardD^{2} - \cardA} \sum_{i, j \notin\mathcal{A}}\text{sim}(\z_i, \z_j)  - \frac{1-\alpha}{\cardA} \sum_{i, j \in\mathcal{A}}\text{sim}(\z_i, \z_j) 
\end{align}
where $\alpha  < 1$ is a hyper-parameter applied in order to compensate for the uncertainty of the anchor pairs being actually dissimilar. 
Let define $\tilde{\z}_i = \frac{\z_i}{\|\z_i\|}$ as the normalized latent space representation. We apply the similarity measure $\text{sim}(\z_i, \z_j) =|\tilde{\z}_i^T\tilde{\z}_j| $.
where $| \cdot |$ stands for the absolute value. Note that eq. \eqref{eq_discrim} is the weigthed $L_1$ norm of all pairwise cosine similarities. We choose the $L_1$ instead of e.g. $L_2$ due to the desired sparsity of the similarities (only $\frac{1}{K}$ non-zero elements) which is encouraged by the $L_1$ norm.

Since generally, the dataset cannot be maintained in main memory, and training is performed on batches of the dataset $\mathcal{B}$ by stochastic gradient descent (SGD), eq. \eqref{eq_discrim} shall be approximated using  the batch matrix of the latent representation $\Z \in R^{|\mathcal{B}| \times d}$, where $|\mathcal{B}|$ denotes the cardinality of the batch. Define $\tilde\Z$ as the row-wise normalized batch matrix (its $i$-th row is the row vector  $\tilde{\z}_i^T$), and $\C = \tilde\Z \tilde\Z^T\in R^{|\mathcal{B}| \times |\mathcal{B}|}$ as the pairwise cosine similarity matrix, such that $C_{i,j} = \tilde{\z}_i^T\tilde{\z}_j$. Furthermore, for each batch a k-nearest neighbors graph is constructed and a set of $\setA_{\setB}$ is determined, yielding the following approximation to eq. \eqref{eq_discrim}:
\begin{align}
\label{eq_discrim_approx}
L_d(\Z; \btheta_e) \approx \frac{1}{|\mathcal{B}| ^2 - |\setA_{\setB}|}\sum_{i,j \notin \setA_{\setB} }|C_{i,j}| - \frac{1-\alpha}{|\setA_{\setB}|}\sum_{i,j \in \setA_{\setB}}C_{i,j}
\end{align}
Note that the diagonal terms of $\C$ are constant and equal $1$, and therefore do not affect the optimization. Furthermore, observe that in the right-hand component we take a sum without absolute value to encourage similarities with value of 1 rather than dissimilar opposite features with value -1.

In order to avoid an arbitrary discrimination of the data-points, we propose to regularize  eq. \eqref{eq_discrim_approx} with  the reconstruction loss, yielding the following optimization problem 
\begin{align}
\label{eq_initial}
\minim{\btheta_e, \btheta_d} \,\,  L_d(\Z; \btheta_e) +\lambda L_r(\X, \hat{\X})
\end{align}  
where $\lambda$ stands for the regularization strength, and $L_r(\X, \hat{\X})=\|\X - \hat{\X} \|_F^2$  denotes the reconstruction loss,  $\X$ stands for the raw input batch matrix and $\|\cdot\|_F$ stands for the Frobenius norm. 

\subsection{The Clustering phase :  maintaining dissimilarity}
After optimizing for $\btheta_e, \btheta_d$ from the pre-training phase, the learned auto-encoder parameters are applied as an initial condition to the clustering phase. In this step we jointly optimize both the encoder-decoder parameters $\btheta_e, \btheta_d$, and the centroids $\centvars \in R^d $;  the new optimization variables for the clustering objective.  A natural candidate for the objective function for the clustering phase is obviously the cosine similarity between the learned centroids and each data-point, since the cosine similarity is applied to discriminate between pairs of data points in the initial phase. Accordingly the primary goal of the clustering phase is to maximize the following objective function
\begin{align}
\label{eq_clustering_obj}
L_c(\Z, \btheta_e, \centvars, \S )= \sum_{i=1}^{\cardB} \sum_{k=1}^K S_{ik}\ \bmu_k^T\tilde{\z}_i \quad s.t. \|\bmu_k \| = 1 \quad \forall\,\, k =1 ,..., K
\end{align}
where $\S$ stands for the assignment matrix and $S_{ik} \in \{0, 1\}$ are the hard decision of the clustering procedure. 
The clustering phase is divided into two steps: in the first step the clustering assignments are not trusted yet, so the clustering objective is kept regularized by eq. \eqref{eq_initial} in which the weights are determined by the anchor pairs. At the second step the clustering assignments are considered reliable and the weights are determined according to the assignments of the clustering. Accordingly the optimization problem solves in the first step is given by
\begin{align}
\label{eq_clustering_1}
\maxim{\btheta_e, \btheta_d, \centvars, \S} L_c(\Z, \btheta_e,\centvars, \S ) - \lambda_d L_d(\Z, \btheta_e) - \lambda_r L_r(\X, \hat{\X})
\end{align}
where $\lambda_d , \lambda_r$ stands for the regularization strength of the discrminative and reconstruction losses,  respectively. For the second step, let define a similarity measure between each pair of clusters, yielding
\begin{align}
\label{eq_between_loss}
L_b(\Z, \btheta_e) = \maxim{k, \ell} \,\,L_{k, \ell}^b(\Z, \btheta_e) ; \quad L_{k, \ell}^b(\Z, \btheta_e) = \frac{1}{\cardCk \cardCl} \sum_{i \in \setCk, j \in \setCl}|\nrmz_i^T\nrmz_j| 
\end{align}
Note that, eq. \eqref{eq_between_loss} penalizes for the worst-case, that is, for the pair of clusters with the greatest similarity. In the same manner we do for the within class objective we have
\begin{align}
\label{eq_within_loss}
L_w(\Z, \btheta_e) = \sum_{k} L_{k}^w(\Z, \btheta_e) ; \quad L_{k}^w(\Z, \btheta_e) = \frac{1}{\cardCk^2} \sum_{i, j \in \setCk}\nrmz_i^T\nrmz_j 
\end{align}

Note that in eq. \eqref{eq_within_loss} the absolute value has been omitted, and the value of $1$ for the cosine similarity between pairs of data-points is preferable over $-1$. The optimization problem in the second step becomes
\begin{align}
\label{eq_clustering_2}
\maxim{\btheta_e, \btheta_d, \{\bmu_k\}, \S} L_c(\Z, \btheta_e, \{\bmu_k\}, \S ) +\lambda_w\, L_w(\Z, \btheta_e)  - \lambda_b L_b(\Z, \btheta_e) - \lambda_r L_r(\X, \hat{\X})
\end{align}
where $\lambda_b, \lambda_w, \lambda_r$ stand for the regularization strength of the between cluster, within cluster and reconstruction loss respectively. 

\subsection{System architecture}
\begin{figure}[t]
  \centering
  \includegraphics[width=0.85\textwidth]{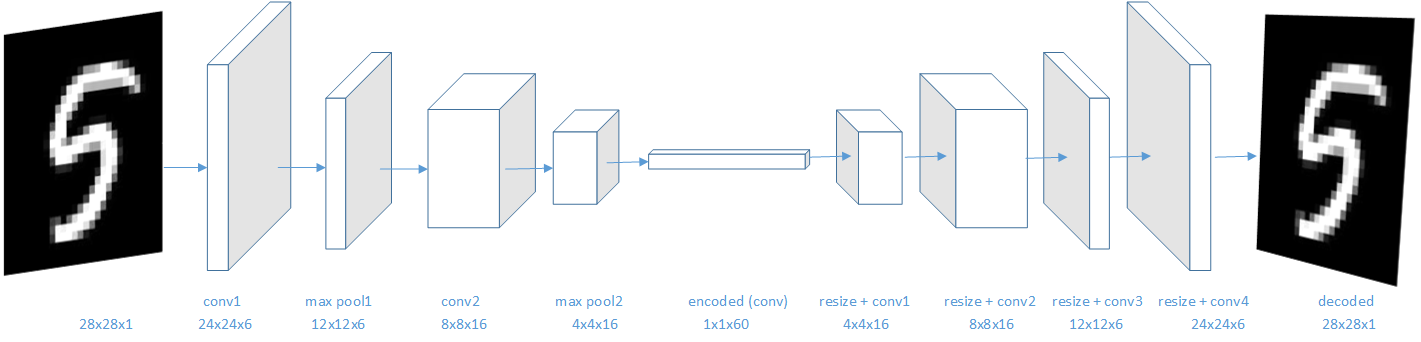}
   \caption{Auto-encoder architecture for MNIST}
    \label{fig_sys_architecture}
\end{figure}

Here we adopt the architecture proposed by \citep{li2017discriminatively} Our system consists of two parts: a
deep discriminative auto-encoder and clustering centroids. The auto-encoder network is a fully convolutional neural network with convolution layers with relu activations followed by a batch normalization \citep{ioffe2015batch} and a max pooling layers. The decoder up-samples the latent space to higher resolution using nearest-neighbor extrapolation that are followed by batch-normed convolution layers as described by \citep{shi2016real}. The auto-encoder architecture is depicted in figure \ref{fig_sys_architecture}.

\subsection{Training strategy}
Training the auto-encoder in the initial phase begins with the minimization of eq. \eqref{eq_initial}. The  regularization strength of the discriminative loss is an hyper-parameter such that $\lambda \in(0, 1]$. The value of $\lambda$  differs among different datasets, such that datasets that are more complex, require more aggressive discrimination while maintaining the strength of the reconstruction loss is constant. The training is done on large batches to ensure $|\setB| \gg K$. As described in \ref{alg_ae} for each batch a k-nearest neighbors graph is constructed, then a small set of anchor pairs is extracted from the graph and eq. \eqref{eq_initial} is backpropagate to optimize for the auto-encoder parameters. The training scheme for the auto encoder is summarized in algorithm \ref{alg_ae}.

\begin{algorithm}[h]
    \caption{Auto-encoder pre-training }
    \label{alg_ae}
    \textbf{Input :} dataset $\setD$, $\alpha,  \lambda,  k \leq K,  \epsilon$  \\
    \textbf{Output :} auto-encoder parameters $\btheta_e, \btheta_d$
    \begin{algorithmic}[1] 
            \While{$i \leq \mathcal{N}_i$ and $|L_d^{i} - L_d^{(i-1)}| > \epsilon$} 
            \State Build: k-nearest neighbors graph $\mathcal{G}_{\setB}$
            \State Extract: anchor pairs $\mathcal{A}_{\setB}\subset \mathcal{G}_{\setB}$
          \State Solve: $ \btheta_e^\star, \btheta_d^\star = \argmin{ \btheta_e, \btheta_d}\,\,   L_d(\Z; \btheta_e) + \lambda L_r(\X, \hat{\X})$ \Comment{Back-propagate for each   $\Z, \X \in \setB$}
         \State $i \leftarrow i + 1$
                \EndWhile
            \State $\z_i = f(\x_i; \btheta_e^\star) \quad \forall \x_i \in \setD$ \Comment{Extract latent space for each data-point  $\x_i \in \setD$}
    \end{algorithmic}
\end{algorithm}

In the clustering phase the clustering variables $\centvars, \S$ are jointly optimized with the auto encoder parameters $\btheta_e, \btheta_d$.  We apply an alternating maximization scheme in which each set of variables is optimized while the other sets remain fixed. The optimization process begins with the initialization of $\centvars$ by optimizing  eq. \eqref{eq_clustering_obj} based on the entire dataset $\setD$. Next we alternate over the maximization of the assignment matrix $\S$, followed by the maximization of $\{\bmu_k\}_{k=1}^K$, and finally maximizing with respect to the auto-encoder parameters. The optimization procedure iterates until convergence.

The clustering phase is divided into two stages that differ from each other by the objective functions they aim to maximize. The pseudo-code for the first stage is summarized in algorithm \ref{alg_clustering_1}.  In the first stage we optimize eq. \eqref{eq_clustering_1}, while using relative large regularization strength for the discriminative loss $\lambda_d \in [1, 5]$ and lower regularization strength $\lambda_r \in (0, 1]$ for the reconstruction loss. 
The while loop refers to the alternating maximization scheme in which each set of parameters is maximized over several epochs. The optimization is carried out using backpropagation for each $\Z, \X \in \setB$. Termination occurs either when the maximal number of iteration $\mathcal{N}_i$ is exceeded or when the clustering objective $L_c$ does not improve over consecutive iterations above a predefined tolerance $\text{tol}$. Note that $\lambda_d, \lambda_r, \lambda_{\min}$ are hyper-parameters and are dataset dependent. 
\begin{algorithm}[h]
    \caption{Clustering phase - stage I }
    \label{alg_clustering_1}
    \textbf{Input :} dataset $\setD,  \btheta_e^\star, \btheta_d^\star, \lambda_d, \epsilon$  \\
    \textbf{Output :} auto-encoder parameters $\btheta_e, \btheta_d$\\
  	 centroids parameters $\centvars$\\
 	 assignments $\S$ \\
	\textbf{Initialize :} $\bmu_k^{(0)} = \argmax{\bmu_k} \sum_{i} S_{ik} \bmu_k^T\tilde{\z}_i \quad \forall \,\,k = 1,..., K$ \Comment{Optimized over the entire dataset $\setD$}
    \begin{algorithmic}[1] 
            \While{$i \leq \mathcal{N}_i$ and $|L_c^{(i)} - L_c^{(i-1)}| \geq \epsilon$} \Comment{ Alternating loop}
            \State $S_{ik}^\star = \left\{  \begin{array}{l l}
1 &\,\,   \tilde{\z}_i^T\bmu_k \geq \tilde{\z}_i^T\bmu_\ell \,\, \forall \ell \neq k\\
0 & \,\, o.w 
\end{array} \right.\ $
            \State  $\bmu_k^\star = \argmax{\bmu_k} \sum_{i} S_{ik} \bmu_k^T\tilde{\z}_i \quad \forall \,\,k = 1,..., K$ \Comment{eq. eq. \eqref{eq_clustering_1} is separable for each $\bmu_k$  }
            \State $ \btheta_e^\star, \btheta_d^\star = \argmax{\btheta_e, \btheta_d}\, L_c(\Z, \btheta_e, \{\bmu_k^\star\}_{k=1}^K, \S^\star ) - \lambda_d L_d(\Z, \btheta_e) - \lambda_r L_r(\X, \hat{\X})$
            \State $i \gets i +1$
            \EndWhile
    \end{algorithmic}
\end{algorithm}

The second stage is initialized with the parameters from the first stage $\btheta_e^\star, \btheta_d^\star, \{\bmu_k^\star\}_{k=1}^K$. Then we optimize for eq. \eqref{eq_clustering_2} wherein, similarly to the previous stage, the discriminative regularization strengths are set to a relative high value, that is, $\lambda_b, \lambda_w \in  [1, 5]$,  while the regularization strength of the reconstruction loss remain unchanged. The process iterates until convergence. The maximization of each set of variables is carried out using backpropagation of large batches over several epochs In both stages we used large batches as in the auto encoder training phase and for several epochs. The entire procedure of the clustering step is similar to the pseudo-code of algorithm \ref{alg_clustering_1} but now with eq. \eqref{eq_clustering_2} and its associated hyper-parameters.

\section{Experiments and results}
\label{results} 
The proposed method has been implemented using Python \texttt{Tensorflow}-1.5, and has been evaluated on four datasets, i.e MNIST handwritten digits dataset \citep{lecun-mnisthandwrittendigit-2010},  COIL20 /COIL100 multi-view object recognition image datasets \citep{Nene96columbiaobject, Nene96objectimage}. Datasets details are presented in table \ref{datasets-table}. 


\begin{table}[h]
  \caption{Datasets statistics}
  \label{datasets-table}
  \centering
  \begin{tabular}{lllll}
    \toprule
    Dataset     & Samples     & Categories  &Image Size   &Channels  \\
    \midrule
    MNIST            & $70$K        & $10$            &$28 \times 28$        & 1 \\
    COIL 20          &$ 1440 $       &  $20$           & $128 \times 128$   & 1   \\
    COIL 100        & $7200$        & $100$          & $128 \times 128$   & 3\\
    \bottomrule
  \end{tabular}
\end{table}

The performance of our method has been examined versus the following baselines: Deep Embedding for clustering (DEC) by \citep{xie2016unsupervised},Deep Clustering Network (DCN) by \citep{yang2016towards}, Deep clustering via joint convolutional autoencoder embedding and relative entropy minimization (DEPICT) by \citep{dizaji2017deep}, Discrimantive Boosted Clustering (DBC) by \citep{li2017discriminatively} (JULE) by  \citep{yang2016joint}, Variational deep embedding (VaDE) by \citep{jiang2016variational}, Neural Clustering -  \citep{saito2017neural},  Deep Continuous Clustering (DCC) \citep{shah2018deep} and Deep Subspace Clustering Networks (DSC-Net) by \citep{ji2017deep}


\subsection{Evaluation Metrics}
Following \citep{xie2016unsupervised}, given the ground truth labels $\{\mathcal{C}_i | 1 \leq i \leq K\}$ and the clustering labels $\{\hat{\mathcal{C}}_i | 1 \leq i \leq K\}$ the clustering performance is evaluated as the accuracy resulting from the optimal one to one mapping $\{m_{ij}\}$ between the ground truth clusters and the resulting clusters. This is carried out by recasting the problem as a linear assignment problem, which is efficiently solved by the Hungarian algorithm \citep{kuhn1955hungarian}. The assignment problem for the clustering accuracy is defined as follows
\begin{align}
\text{ACC} = \maxim{m_{ij}} \frac{\sum_{i,j } m_{ij}|\mathcal{C}_i  \cap \hat{\mathcal{C}}_j  |}{\cardD}\nonumber
\end{align} 
where $|\mathcal{C}_i  \cap \hat{\mathcal{C}}_j  |$ stands for the cardinality of the intersection of $\mathcal{C}_i,  \hat{\mathcal{C}}_j$, i.e. the number of mutual members of both sets and the mapping variables  $\{m_{ij}\}$ are given by
\begin{align}
\label{eq_ls_mapping}
m_{ij} = \left\{  \begin{array}{l l} \nonumber
1 &\,\,   \hat{\mathcal{C}}_j \mapsto  \mathcal{C}_i\\
0 & \,\, o.w 
\end{array} \right.\
\end{align}
such that both $\sum_i m_{ij} = 1$ and $\sum_j m_{ij} = 1$ are satisfied, which establishes the one to one mapping.

%
%

\subsection{MNIST Dataset}
\begin{figure*}[t]
        \centering
        \captionsetup{justification=centering}
        \begin{minipage}[t]{0.48\textwidth}
               \fbox{ \includegraphics[width=\textwidth]{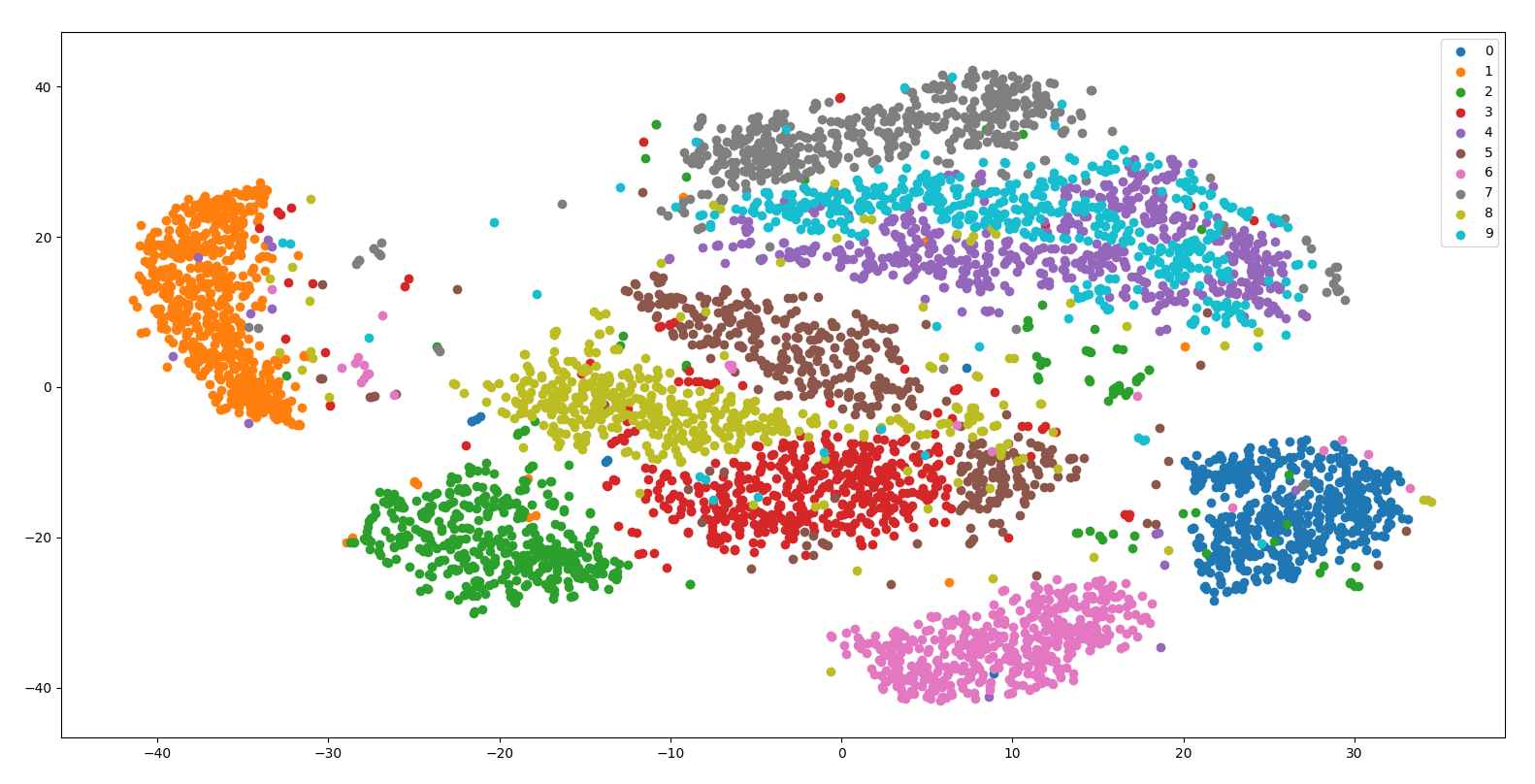}}
                \caption{Raw data - 54\% accuracy}
                \label{fig_raw_data_54}
        \end{minipage}%
        \quad 
        \begin{minipage}[t]{0.43\textwidth}
                \fbox{\includegraphics[width=\textwidth]{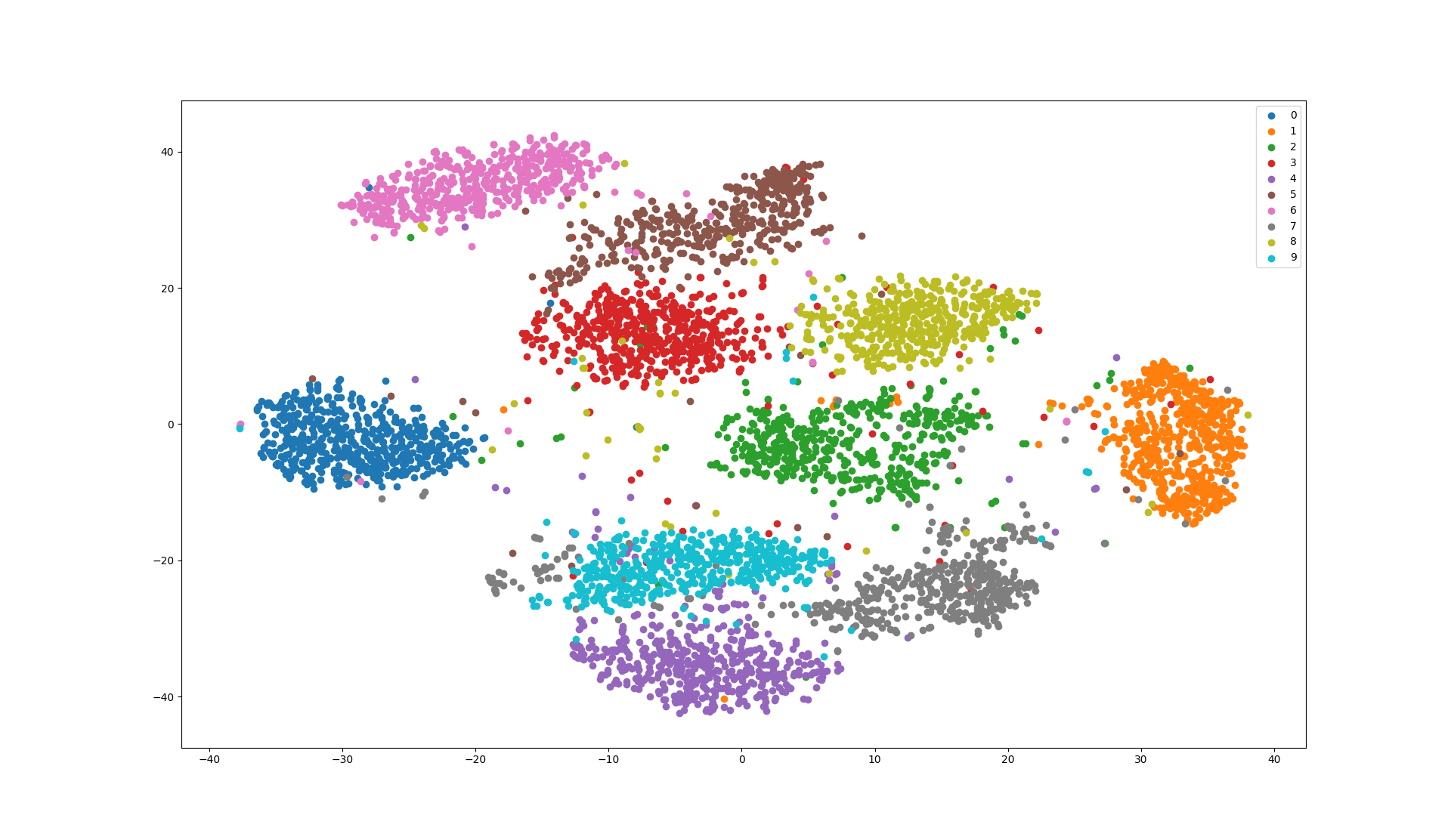}}
                \caption{Encoder phase - 30 Epoch 87\% accuracy}
                \label{fig_auto_encoder_85}
        \end{minipage}
         \quad 
        \begin{minipage}[t]{0.43\textwidth}
                \fbox{\includegraphics[width=\textwidth]{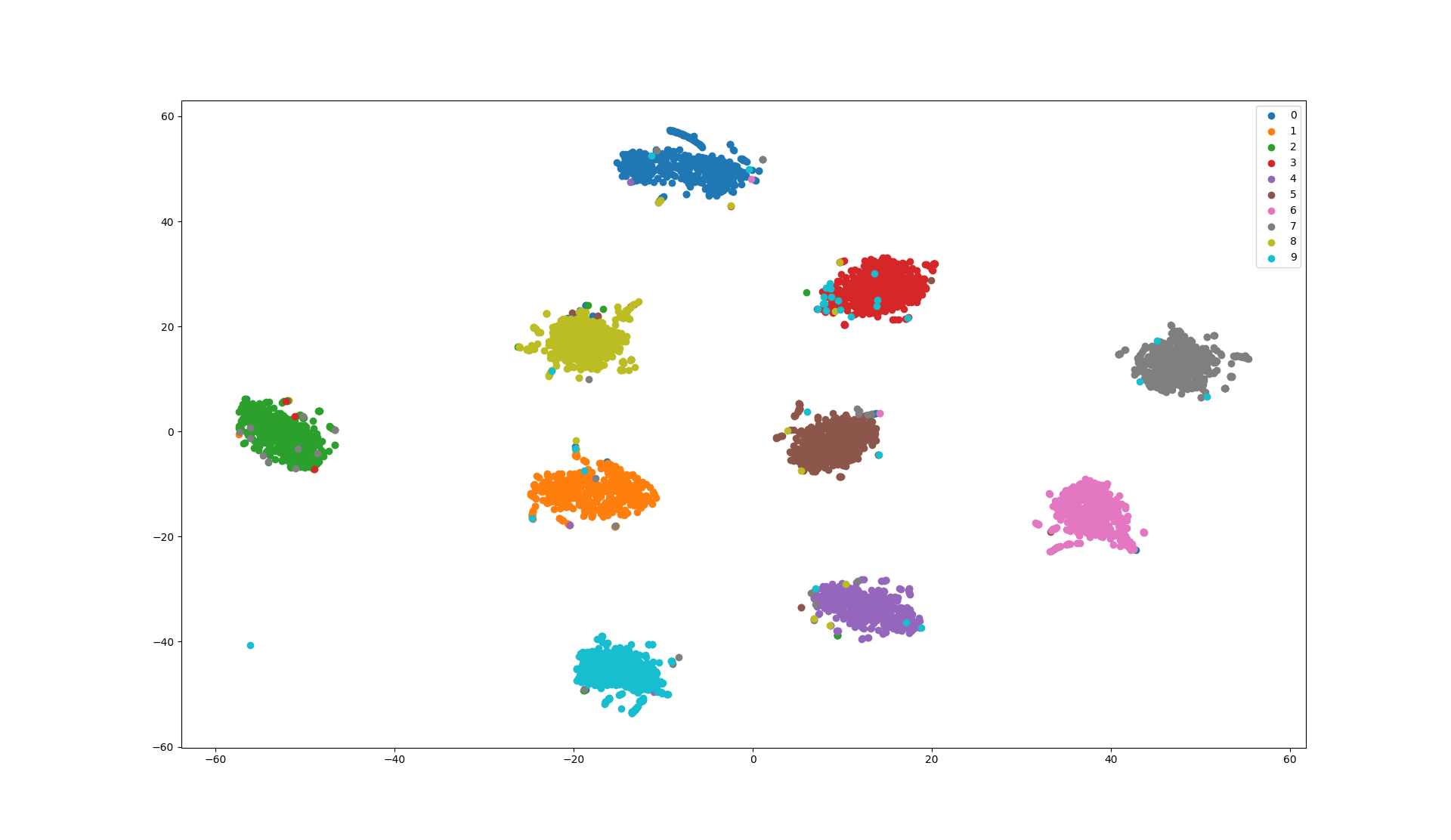}}
                \caption{Clustering Phase Stage II -  50 Epochs 97.4\% accuracy}
                \label{fig_clustering_97}
        \end{minipage}
              \caption{Latent space representation of the auto-encoder $\Z = f(\X, \btheta^\star_e)$  at different phases of the algorithm, represented by t-SNE on $10$K data-points from 
MNIST dataset, and the clustering accuracy achieved by our proposed deep clustering method  }
              \label{fig_mnist_tsne}
\end{figure*}

The MNIST dataset constitutes of $70$K samples of gray scale $28 \times 28$ hand-written digits images, distributed over $10$ categories ($0-9$). For MNIST we used a $3$-layer encoder followed by a $3$-layer decoder, in which the bottleneck of the encoder, i.e. the latent space dimension has been set to $d=60$. 
The overall number of parameters of the entire network is $\sim 3.2$K. 
At the initial phase; the auto-encoder pre-training, the  regularization strength of the reconstruction loss in eq. \eqref{eq_initial} $\lambda$ has been set to $0.001$. The auto-encoder reaches $\text{ACC} \geq 92\%$ prior to the clustering step in less than $50$ epochs. We use relatively large batches with $|\setB| \geq 1000$ to maintain the distribution of the data as in the original dataset such that the ratio between the within cluster pairs and between cluster pairs:
\begin{align}
 \frac{\mathcal{N}_w \in \setD}{\cardD} \approx \frac{\mathcal{N}_w \in \setB}{\cardB} \approx \frac{1}{K} \nonumber
 \end{align}
is preserved, and therefore the assumption that $\C$ in eq. \eqref{eq_discrim_approx} is sparse is justified. 
The clustering phase has been initialized with the auto-encoder parameters obtained from the initial phase. At the first stage of the clustering process, the regularization strength of the discriminative objective, $\lambda$ has been set to $1$. Similarly to the pre-training phase we used large batches jointly optimize for the parameters of the centroids, auto-encoder and the assignment matrix according to the alternating scheme describe in algorithm \ref{alg_clustering_1}. At the second stage of the clustering process the regularization strength of the between cluster similarity and within cluster similarity  in eq. \eqref{eq_clustering_2} have been set to $\lambda_b = 1, \lambda_w=0.3$, respectively. The separability of the latent space of MNIST represented using t-SNE is depicted in figure \ref{fig_mnist_tsne} during the different phases of the algorithm. The algorithm reaches accuracy of $97.4\%$ on both the train and the test sets at the final stage.

\subsection{COIL-20 and COIL-100 Datasets}
We examined our method on the COIL multi-view object image datasets, COIL-20 and COIL-100 by \citep{Nene96columbiaobject, Nene96objectimage}. COIL-20 consists of $1440$  gray-scale $128 \times 128$ image samples distributed over $20$ objects. Similarly, COIL-100 consists of $7200$ colored $128 \times 128$ images distributed over $100$ categories. For both datasets there are $72$ images for each category taken at pose intervals of $5$ degrees. The Images were down-sampled to $32 \times 32$. We apply the same architecture that applied to MNIST with a bottleneck dimension of the encoder $\z$ was  increased to $d =150, 500$ for COIL-20 and COIL-100, respectively. At the initial phase; the  regularization strength of the reconstruction loss  eq. \eqref{eq_initial} $\lambda$ has been set to $0.001$ . At both phases of the algorithm we use a batch size of $\cardB$ of $500$ and $1000$ for COIL-20, and COIL-100, respectively. Since COIL-20 is a small dataset, i.e. (only $1140$ samples in the training set) we utilize relatively small batch size in order to maintain a diversity of anchor pairs across batches. The auto-encoder reaches $\text{ACC} \geq 90\%$ and $89\%$ prior to the clustering step on COIL-20 and COIL 100, respectively. At the first stage of the clustering process, the regularization strength of the discriminative function $\lambda$ has been set to $4$ and decreased to $1$ during the training process. The algorithm reaches accuracy of $92.3\%$ and $90.3\%$ on COIL-20 and COIL-100 at the final stage, respectively. Note that DSC-net reaches accuracy of $94.6\%$ on COIL-20, however this is achieved with a network of more than $1$M parameters, that is quadratic in the size of the dataset, which is obviously not scalable, whereas our network consists of $\sim 7.5$K and $\sim 12.5$K parameters for COIL-20 and COIL -100, respectively (the difference arises due to the different bottleneck size). Observe that on COIL-100  JULE reaches higher accuracy than our method, that is, $91.1\%$ versus $90.3\%$ achieved by the proposed method.


\begin{table}[h]
  \caption{Clustering accuracy ($\%$) performance comparison on all datasets}
  \label{results_table}
  \centering
  \begin{tabular}{llll}
    \toprule
    Method                         & MNIST         & COIL 20  & COIL 100     \\
    \midrule
    DEC \footnote{The authors have not reported performance on COIL 20 
  the numbers here are based on the reports of the authors of DCC- \citep{shah2018deep}}                             & $84.3$        & --            &$81.5$         \\
    DCN\footnotemark[2]                             &$ 83. $          & --            & $62$    \\
    DEPICT\footnotemark[2]                        & $96.5$        & $42.$          & $66.7$  \\
    DBC                             & $96.4$         & $79.3$            & $77.5$       \\
    JULE\footnote{The authors have not reported performance in terms of accuracy but only in terms of NMI 
  the numbers here are based on the reports of the authors of DCC- \citep{shah2018deep}}                           & $80.$           & --            & $\bm{91.1}$      \\
    VaDe                           & $94.4$        & --            & --       \\
    Neural Clustering          & $96.6$        & --            & --       \\
    DSC-Net        		      & --               & $\bm{94.6}$       & $69.1$       \\
    DCC                	      & $96.3$       & --                 & $85.8$      \\
    \textbf{Proposed Method}         & $\bm{97.4}$  & $\bm{92.3}$           & $\bm{90.3}$      \\
    \bottomrule
  \end{tabular}
\end{table}

%
%
%
%

\section{Conclusions}
\label{conclusions}
In this paper we proposed an efficient method for learning a latent space representation for clustering. We propose to minimize a pairwise discriminative function - the weighted sum of all pairwise similarities within the batch matrix with respect to the auto-encoder parameters prior to the clustering phase. We demonstrate the higher accuracy and rapid convergence it achieves, as well as the small models it can handle. 
However, it seems that there is an inherent limit in clustering arbitrary datasets. There are many ways of separating or grouping the dataset and there is no guaranty that the separation obtained by a clustering process will coincide with the ground truth-labels. Prior knowledge about small fraction of the dataset is a reasonable assumption in many applications and utilizing this knowledge within the deep clustering approach may lead to a much superior performance, similar to the achievements of one-shot learning. 
\medskip

%
\bibliography{references}

\begin{thebibliography}{17}
\providecommand{\natexlab}[1]{#1}
\providecommand{\url}[1]{\texttt{#1}}
\expandafter\ifx\csname urlstyle\endcsname\relax
  \providecommand{\doi}[1]{doi: #1}\else
  \providecommand{\doi}{doi: \begingroup \urlstyle{rm}\Url}\fi

\bibitem[Chi and Lange(2015)]{chi2015splitting}
Eric~C Chi and Kenneth Lange.
\newblock Splitting methods for convex clustering.
\newblock \emph{Journal of Computational and Graphical Statistics}, 24\penalty0
  (4):\penalty0 994--1013, 2015.

\bibitem[Dizaji et~al.(2017)Dizaji, Herandi, Deng, Cai, and
  Huang]{dizaji2017deep}
Kamran~Ghasedi Dizaji, Amirhossein Herandi, Cheng Deng, Weidong Cai, and Heng
  Huang.
\newblock Deep clustering via joint convolutional autoencoder embedding and
  relative entropy minimization.
\newblock In \emph{2017 IEEE International Conference on Computer Vision
  (ICCV)}, pages 5747--5756. IEEE, 2017.

\bibitem[Ioffe and Szegedy(2015)]{ioffe2015batch}
Sergey Ioffe and Christian Szegedy.
\newblock Batch normalization: Accelerating deep network training by reducing
  internal covariate shift.
\newblock \emph{arXiv preprint arXiv:1502.03167}, 2015.

\bibitem[Ji et~al.(2017)Ji, Zhang, Li, Salzmann, and Reid]{ji2017deep}
Pan Ji, Tong Zhang, Hongdong Li, Mathieu Salzmann, and Ian Reid.
\newblock Deep subspace clustering networks.
\newblock In \emph{Advances in Neural Information Processing Systems}, pages
  23--32, 2017.

\bibitem[Jiang et~al.(2016)Jiang, Zheng, Tan, Tang, and
  Zhou]{jiang2016variational}
Zhuxi Jiang, Yin Zheng, Huachun Tan, Bangsheng Tang, and Hanning Zhou.
\newblock Variational deep embedding: An unsupervised and generative approach
  to clustering.
\newblock \emph{arXiv preprint arXiv:1611.05148}, 2016.

\bibitem[Kuhn(1955)]{kuhn1955hungarian}
Harold~W Kuhn.
\newblock The hungarian method for the assignment problem.
\newblock \emph{Naval Research Logistics (NRL)}, 2\penalty0 (1-2):\penalty0
  83--97, 1955.

\bibitem[LeCun and Cortes(2010)]{lecun-mnisthandwrittendigit-2010}
Yann LeCun and Corinna Cortes.
\newblock {MNIST} handwritten digit database.
\newblock 2010.
\newblock URL \url{http://yann.lecun.com/exdb/mnist/}.

\bibitem[Li et~al.(2017)Li, Qiao, Zhang, and Xi]{li2017discriminatively}
Fengfu Li, Hong Qiao, Bo~Zhang, and Xuanyang Xi.
\newblock Discriminatively boosted image clustering with fully convolutional
  auto-encoders.
\newblock \emph{arXiv preprint arXiv:1703.07980}, 2017.

\bibitem[Nene et~al.(1996{\natexlab{a}})Nene, Nayar, and
  Murase]{Nene96columbiaobject}
Sameer~A. Nene, Shree~K. Nayar, and Hiroshi Murase.
\newblock Columbia object image library (coil-20.
\newblock Technical report, 1996{\natexlab{a}}.

\bibitem[Nene et~al.(1996{\natexlab{b}})Nene, Nayar, and
  Murase]{Nene96objectimage}
Sameer~A. Nene, Shree~K. Nayar, and Hiroshi Murase.
\newblock object image library (coil-100.
\newblock Technical report, 1996{\natexlab{b}}.

\bibitem[Saito and Tan(2017)]{saito2017neural}
Sean Saito and Robby~T Tan.
\newblock Neural clustering: Concatenating layers for better projections.
\newblock 2017.

\bibitem[Shah and Koltun(2017)]{shah2017robust}
Sohil~Atul Shah and Vladlen Koltun.
\newblock Robust continuous clustering.
\newblock \emph{Proceedings of the National Academy of Sciences}, 114\penalty0
  (37):\penalty0 9814--9819, 2017.

\bibitem[Shah and Koltun(2018)]{shah2018deep}
Sohil~Atul Shah and Vladlen Koltun.
\newblock Deep continuous clustering.
\newblock \emph{arXiv preprint arXiv:1803.01449}, 2018.

\bibitem[Shi et~al.(2016)Shi, Caballero, Husz{\'a}r, Totz, Aitken, Bishop,
  Rueckert, and Wang]{shi2016real}
Wenzhe Shi, Jose Caballero, Ferenc Husz{\'a}r, Johannes Totz, Andrew~P Aitken,
  Rob Bishop, Daniel Rueckert, and Zehan Wang.
\newblock Real-time single image and video super-resolution using an efficient
  sub-pixel convolutional neural network.
\newblock In \emph{Proceedings of the IEEE Conference on Computer Vision and
  Pattern Recognition}, pages 1874--1883, 2016.

\bibitem[Xie et~al.(2016)Xie, Girshick, and Farhadi]{xie2016unsupervised}
Junyuan Xie, Ross Girshick, and Ali Farhadi.
\newblock Unsupervised deep embedding for clustering analysis.
\newblock In \emph{International conference on machine learning}, pages
  478--487, 2016.

\bibitem[Yang et~al.(2016{\natexlab{a}})Yang, Fu, Sidiropoulos, and
  Hong]{yang2016towards}
Bo~Yang, Xiao Fu, Nicholas~D Sidiropoulos, and Mingyi Hong.
\newblock Towards k-means-friendly spaces: Simultaneous deep learning and
  clustering.
\newblock \emph{arXiv preprint arXiv:1610.04794}, 2016{\natexlab{a}}.

\bibitem[Yang et~al.(2016{\natexlab{b}})Yang, Parikh, and Batra]{yang2016joint}
Jianwei Yang, Devi Parikh, and Dhruv Batra.
\newblock Joint unsupervised learning of deep representations and image
  clusters.
\newblock In \emph{Proceedings of the IEEE Conference on Computer Vision and
  Pattern Recognition}, pages 5147--5156, 2016{\natexlab{b}}.

\end{thebibliography}

%
%

\end{document}